\documentclass{article}

\usepackage[nonatbib,final]{neurips_data_2022}

\usepackage[utf8]{inputenc} 
\usepackage[T1]{fontenc}    
\usepackage{hyperref}       
\usepackage{url}            
\usepackage{booktabs}       
\usepackage{amsfonts}       
\usepackage{nicefrac}       
\usepackage{microtype}      
\usepackage[table,xcdraw,dvipsnames]{xcolor}         

\usepackage{float}
\usepackage{multicol}
\usepackage{multirow}
\usepackage{amssymb}
\usepackage{wrapfig}
\usepackage{graphicx}
\usepackage{diagbox}
\usepackage{ulem}
\usepackage{color}
\definecolor{linkcolor}{RGB}{255,0,0}

\usepackage{hyperref}
\hypersetup{colorlinks=true,citecolor=green,linkcolor=linkcolor,urlcolor=Thistle}

\usepackage{xspace}
\def\onedot{.\xspace}
\def\eg{\normalem{\emph{e.g}\onedot}} 

\def\ie{\normalem\emph{i.e}\onedot}

\def\etc{\normalem{\emph{etc}\onedot}}

\title{APT-36K: A Large-scale Benchmark for Animal Pose Estimation and Tracking}

\makeatletter
\renewcommand*{\@fnsymbol}[1]{\ensuremath{\ifcase#1\or \dagger\or \ddagger\or
   \mathsection\or \mathparagraph\or \|\or **\or \dagger\dagger
   \or \ddagger\ddagger \else\@ctrerr\fi}}
\makeatother

\author{%
  Yuxiang Yang$^1$, Junjie Yang$^{1}$, Yufei Xu$^{2}$\thanks{Corresponding authors.}\ , Jing Zhang$^{2\dag}$, Long Lan$^{3}$, Dacheng Tao$^{4}$\\
$^1$ School of Electronics and Information, Hangzhou Dianzi University, Hangzhou 310018, China \\
$^2$ School of Computer Science, The University of Sydney, NSW 2006, Australia\\
$^3$ National University of Defense Technology, Changsha 410073, China\\
$^4$ JD Explore Academy, Beijing 101111, China\\
\texttt{yyx@hdu.edu.cn, 212040105@hdu.edu.cn, yuxu7116@uni.sydney.edu.au}\\
  \texttt{jing.zhang1@sydney.edu.au, long.lan@nudt.edu.cn, dacheng.tao@gmail.com}\\
}
\begin{document}

\maketitle

\begin{abstract}

Animal pose estimation and tracking (APT) is a fundamental task for detecting and tracking animal keypoints from a sequence of video frames. Previous animal-related datasets focus either on animal tracking or single-frame animal pose estimation, and never on both aspects. The lack of APT datasets hinders the development and evaluation of video-based animal pose estimation and tracking methods, limiting real-world applications, e.g., understanding animal behavior in wildlife conservation. To fill this gap, we make the first step and propose APT-36K, i.e., the first large-scale benchmark for animal pose estimation and tracking. Specifically, APT-36K consists of 2,400 video clips collected and filtered from 30 animal species with 15 frames for each video, resulting in 36,000 frames in total. After manual annotation and careful double-check, high-quality keypoint and tracking annotations are provided for all the animal instances. Based on APT-36K, we benchmark several representative models on the following three tracks: (1) supervised animal pose estimation on a single frame under intra- and inter-domain transfer learning settings, (2) inter-species domain generalization test for unseen animals, and (3) animal pose estimation with animal tracking. Based on the experimental results, we gain some empirical insights and show that APT-36K provides a valuable animal pose estimation and tracking benchmark, offering new challenges and opportunities for future research. The code and dataset will be made publicly available at \href{https://github.com/pandorgan/APT}{https://github.com/pandorgan/APT-36K}.

\end{abstract}

\section{Introduction}
\label{sec:introduction}

\begin{figure}[htbp]
    \center
    \includegraphics[width=\textwidth]{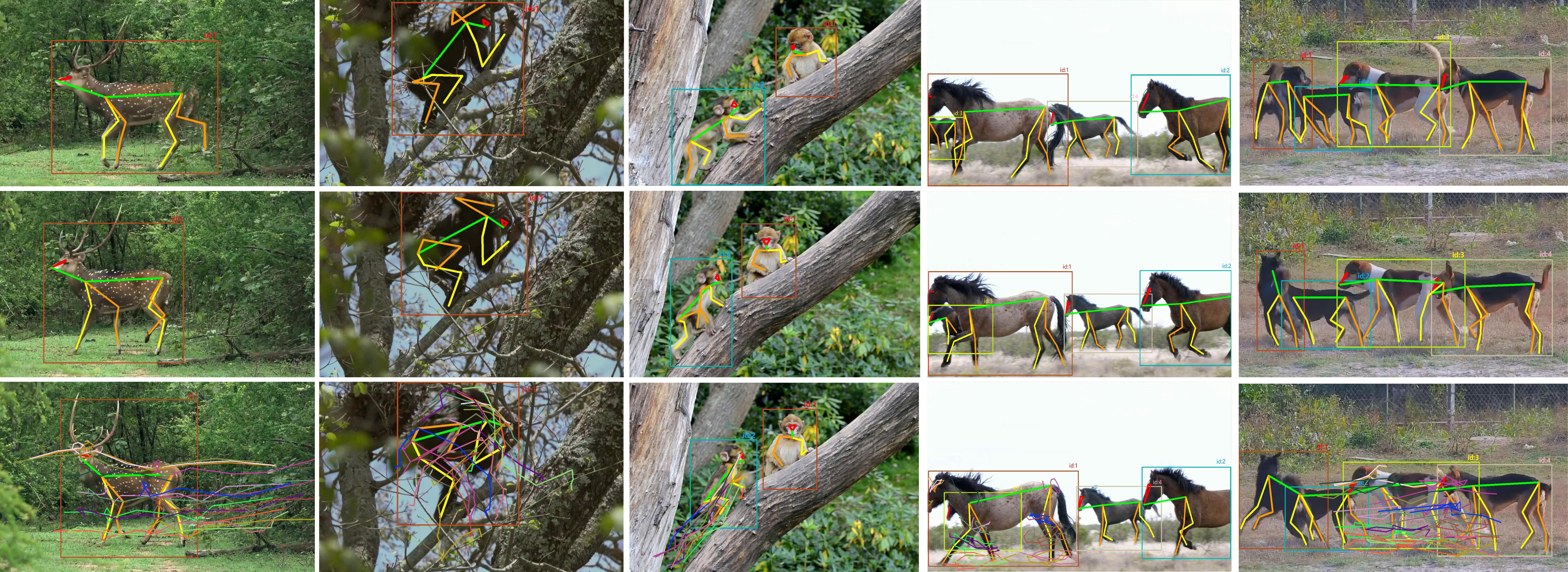}
    \caption{A glance of some examples from the proposed APT-36K dataset. The five columns of images show single animal instance, complex movements, and multiple animal instances, respectively. The keypoint trajectories of some animal instances are shown in the last frames. Best viewed in color.}
    \label{fig:demo}
\end{figure}

Pose estimation aims to identify the categories and distinguish the locations of a series of body keypoints from an image. As a fundamental task in computer vision, it is beneficial for many vision tasks~\cite{ni2017learning,baradel2018glimpse,arac2019deepbehavior,zhang2020empowering} like behavior understanding, action recognition, etc. There has been rapid progress in human pose estimation thanks to the availability of a large number of human pose datasets~\cite{lin2014microsoft,li2018crowdpose,mpii}. However, fewer works focus on animal pose estimation, especially for video-based animal pose estimation, although it is crucial in animal behavior understanding and wildlife conservation.

Some efforts have been made to establish animal pose estimation datasets to facilitate research in this area. Early works focus on the pose estimation of specific animal categories, \eg, the horse~\cite{mathis2021pretraining}, zebra~\cite{graving2019deepposekit}, macaque~\cite{labuguen2020macaquepose}, fly~\cite{pereira2019fast}, and tiger~\cite{tiger} datasets collect and annotate the keypoints for some specific kind of animals. They do help advance the research of pose estimation for these animals. However, since there exist huge appearance variance, behavior difference, and joint distribution shifts among various animal species due to evolution, the models trained on these datasets do not perform well on unseen animal species, leading to poor generalization performance. To further facilitate the research on animal pose estimation, datasets covering multiple animal species with keypoint annotations are proposed, \eg, Animal-Pose~\cite{Cao_2019_ICCV} and AP-10K~\cite{yu2021ap}. Despite their large scale in dataset volume and diversity in animal species, they lack important temporal information, making it unexplored to recognize poses in contiguous frames, which is very important for animal action recognition and beyond. 

To fill this gap, we propose APT-36K, i.e., the first large-scale dataset with high-quality animal pose annotations from consecutive frames for animal pose estimation and tracking. APT-36K consists of 2,400 video clips collected and filtered from 30 different animal species with 15 frames for each video, resulting in 36,000 frames in total. These animals can be further classified into 15 animal families following taxonomic rank to facilitate the evaluation of inter-species and inter-family generalization ability of pose estimation models. Specifically, the video clips are collected from YouTube videos with carefully filtering. Then, the frames are sampled with specific intervals (i.e., at a low frame-per-second (FPS) speed) to remove duplication and increase the temporal motion amplitude. Next, 16 well-trained annotators are recruited to label the keypoints for each animal in each frame following the labeling routines of MS COCO~\cite{lin2014microsoft}, which are then manually double-checked. The trajectory of each animal is also denoted with bounding boxes and exclusive instance ids across the videos. In this way, APT-36K can support the research on both single-frame pose estimation and animal estimation tracking in consecutive frames.

Based on APT-36K, we set up three tracks to benchmark previous state-of-the-art (SOTA) pose estimation methods~\cite{sun2019deep,xiao2018simple,yuan2021hrformer,xu2022vitpose,xu2021vitae}, \ie, (1) single-frame animal pose estimation (SF track), (2) inter-species domain generalization (IS track), and (3) animal pose tracking (APT track). In the basic SF track, we comprehensively evaluate the performance of representative convolutional neural networks (CNN) and vison transformer-based methods in different settings, including inter- and intra-domain transfer learning, where the models are pre-trained on ImageNet dataset~\cite{deng2009imagenet}, MS COCO human pose estimation dataset~\cite{lin2014microsoft}, and AP-10k animal pose estimation dataset~\cite{yu2021ap}, respectively. In the IS track, the inter-family domain generalization ability of different pose estimation methods is evaluated, where the model is trained on images of all species from a certain family and tested on the images of other families. 
In the APT track, several object trackers, including a customized one based on the plain vision transformer~\cite{dosovitskiy2020image}, are used to track the animal instances, and representative pose estimation methods are used to detect the keypoints of the tracked instances, where their performance is evaluated accordingly. The detailed experiment settings and results are presented in Section~\ref{sec:experiment}, from which we demonstrate that the great potential of vision transformers in both animal pose estimation and animal pose tracking, the benefits of knowledge transfer between human pose estimation and animal pose estimation, as well as the advantages of employing diverse animal species for animal pose estimation.

The main contribution of this paper is two-fold. First, we establish the first large-scale benchmark APT-36K for animal pose estimation and tracking. Its large scale, diversity of animal species, and abundant annotations of keypoints, bounding boxes, and instance ids among consecutive frames make it a good test bed for future study. Second, we set up three challenging tasks, including SF, IS, and APT, based on APT-36K and comprehensively benchmark representative pose estimation methods built upon both CNN and vision transformers, gaining some useful insights.

\section{Related work}
\subsection{Human pose estimation}
Pose estimation is a fundamental computer vision task for many applications such as behavior understanding. In the past decades, significant progress has been made in human pose estimation no matter in datasets~\cite{lin2014microsoft,li2018crowdpose,mpii} and methods~\cite{xiao2018simple,sun2019deep,yuan2021hrformer,xu2022vitpose,zhang2021towards}. For example, MPII~\cite{mpii} and MS COCO~\cite{lin2014microsoft} are two popular large-scale benchmarks for human pose estimation. To further evaluate the performance of human pose estimation methods regarding challenging scenarios such as occlusion or crowd, OCHuman~\cite{ochuman} and CrowdPose~\cite{li2018crowdpose} are established where there are heavy occlusions of body keypoints or multiple human instances in a single frame. Both top-down and bottom-up methods have been proposed and evaluated based on these datasets. Despite the significant contribution made by these works, the temporal information 
has been largely ignored, which is important to understanding human behavior and action imitation from humans to robots. To address this issue, several video-based pose estimation datasets have been proposed, \eg, VideoPose~\cite{videopose}, YouTube Pose~\cite{charles2016personalizing}, J-HMDB~\cite{jhuang2013towards}, and PoseTrack~\cite{andriluka2018posetrack}. They facilitate the research of human pose estimation and tracking~\cite{xiao2018simple}. 

\subsection{Animal pose estimation}
Recently, animal pose estimation has attracted increasing attention from the research community due to animal behavior understanding and wildlife conservation demand. Generally, animal pose estimation methods share similar ideas to human pose estimation ones, \eg, bottom-up and top-down methods based on heatmap regression~\cite{yu2021ap}. Nevertheless, different from the human pose, where the appearance, movement pattern, and keypoint distribution are similar among different people, they vary significantly for different animal species due to the difference in their habitat and evolutionary route. 
To facilitate the research in this area, many datasets have been proposed. 
In early works, single category animal pose estimation datasets are introduced, \eg, horse~\cite{mathis2021pretraining}, zebra~\cite{graving2019deepposekit}, macaque~\cite{labuguen2020macaquepose}, fly~\cite{pereira2019fast}, and tiger~\cite{tiger} datasets. However, models trained on these datasets suffer from a limited generalization ability due to the significant difference in appearance and movement pattern between different animal species. To address this issue, some datasets covering many animal species have been established, including Animal-Pose~\cite{Cao_2019_ICCV}, Animal Kingdom~\cite{ng2022animal}, and AP-10K~\cite{yu2021ap}. For example, AP-10K contains 10,015 annotated images from 23 animal families and 54 species. Nevertheless, there are no temporal annotations in these datasets, making it impossible to develop animal pose tracking methods. Recently, a dataset named AnimalTrack for animal tracking has been established~\cite{zhang2022animaltrack}. However, it only focuses on animal instance tracking rather than the fine-grained keypoint tracking. Besides, it has only limited video clips and animal species, \eg, fewer than 60 video clips and 10 animal categories. Different from the above works, we propose APT-36K to fill the gap of the lack of real-world animal pose tracking datasets. Thanks to its large scale, diversity of animal species, and abundant annotations of keypoints, bounding boxes, and instance ids, we believe our APT-36K will benefit the research of animal pose estimation and tracking by serving as a training data source as well as a test bed along with several well-defined benchmark tracks. 

\subsection{Visual object tracking}
Object tracking~\cite{li2019siamrpn,ni2017learning,hong2015multi,kristan2015visual} is a fundamental and active research topic in computer vision. One popular direction for object tracking follows the tracking by detection routine. For example, given the current frame and subsequent frames, an object detector is first used to detect the candidates from subsequent frames. Then, different techniques are employed to associate the detection results with the target in the current frame. These methods obtain superior results in both multiple object tracking (MOT)~\cite{wojke2017simple,zhang2021bytetrack} and single object tracking (SOT)~\cite{bertinetto2016fully,zhang2020ocean,wang2021transformer}. However, the generalization abilities of these trackers are limited, \ie, the objects they can track should belong to the categories that the detectors support. The limitation hinders their usage in animal pose tracking, where many animal species may be unseen during the training of the detectors. The other development direction of object tracking follows the tracking by matching pipeline, \ie, a siamese network is utilized to extract features from the tracked targets in the previous frame and the search regions in the subsequent frame, and then the two kinds of features are matched to localize the targets in the search region. In this paper, we mainly adopt this kind of tracking method for animal instance tracking since they do not make assumptions about the target categories. Besides, we benchmark their performance for animal pose tracking by combining them with different animal pose estimation methods. 

\section{Dataset}
In this section, we briefly introduce the details of the proposed APT-36K dataset, including data collection and organization, data annotation, and the statistics of the dataset. Moreover, we provide detailed datasheets and more results in the supplementary material.

\subsection{Data collection and organization}
The goal of APT-36K is to provide a large-scale benchmark for animal pose estimation and tracking in real-world scenarios, which has been rarely explored in prior art. 
To this end, we resort to real-world video websites, \ie, YouTube, and carefully collect and filter 2,400 video clips covering 30 different animal species from different scenes, \eg, zoo, forest, and desert. However, directly annotating these videos and using them as training data is not appropriate since the movement speed of different animals and the frame frequency of different videos vary a lot, \eg, some animals are almost static during a specific period. To address this issue, we manually set the frame sampling rate for each video to ensure there are noticeable movement and posture differences for each animal in the sub-sampled video clips. Specifically, each clip contains 15 frames after the sampling process. It should be noted that challenging cases such as truncation and occlusion are kept in the dataset owing to the above process, making it possible to evaluate the models regarding these challenges. 

After the video collection and cleaning process, we categorize the videos from 30 animal species further into 15 families following the taxonomic rank. Following the terms of YouTube, we use sparsely sampled frames in the videos to formulate the APT-36K dataset and use them for research purposes only. According to Linnean's theory of evolution, animals belonging to the same taxonomic rank may share more similarities in behavior patterns, anatomical keypoint distribution, and appearance than those belonging to different families. For example, the walking posture of dogs and wolves is similar since they belong to the Canidae family, while zebras' walking patterns are far from similar to them since it belongs to a different family, \ie, Equidae. Following the taxonomic rank, the proposed dataset can be easily scaled up by collecting and annotating more animal images from the same species or families, as well as different ones. Moreover, it also implies that such an organized way of the animal pose dataset provides a possible way to enhance the generalization ability of animal pose estimation models to rare animal species, \ie, by collecting and annotating images from other more common animals of the same taxonomic rank.

\subsection{Data annotation}

To guarantee high-quality annotations for each image in the APT-36K dataset, 16 well-trained annotators participated in the annotation process, and one strict cross-check was then carried out to improve the annotation quality. The annotation-check round is repeated three times during the labeling process. The whole data collection, cleaning, annotation, and check process takes about 2,000 person-hours. A total of 36,000 images are finally labeled, following the COCO labeling format. There are typically 17 keypoints labeled for each animal instance, including two eyes, one nose, one neck, one tail, two shoulders, two elbows, two knees, two hips, and four paws as~\cite{yu2021ap}. It should be noted that we do not exactly follow the biological definition to localize the keypoints for specific animals, \ie, we use the paw to define the tire point of horses and the knee to represent the end of their hock. In this way, it helps us better figure out the behavior of specific animals since half of the horses' legs will have no annotations if we strictly follow the anatomy definition. Besides the keypoint annotations, we label the background type for each frame from 10 classes, \ie, grass, city, and forest. In addition, we label each same animal instance across the video clips with a unique tracking id. The annotations are also manually checked for two rounds to improve their quality. The dataset is split into three disjoint subsets for training, validation, and test, respectively, following the ratio of 7:1:2 per animal species. It is also noteworthy that we adopt a video-level partition to prevent the potential information leakage since the frames in the same video clip are similar to each other.

\subsection{Statistics of the APT-36K dataset}

\begin{table}[htbp]
  \centering
  \scriptsize
  \caption{Comparison of different animal pose datasets.}
    \setlength{\tabcolsep}{0.003\linewidth}{\begin{tabular}{c|ccccccc}
    \hline
          & {\#Species} & {\#Family} & {\#Labeled image} & {\#Keypoint} & {\#Sequence} & {\#Instance} &{\#Background type} \\
    \hline
    Horses-10~\cite{mathis2021pretraining}  & 1 & 1 & 8,100 & 22 & N/A & 8,110 & N/A \\
    Animal-Pose Dataset~\cite{Cao_2019_ICCV}  & 5 & N/A & 4,666 & 20 & N/A & 6,117 & N/A \\
    Animal kingdom~\cite{ng2022animal}  & 850 & 6 & 33,099 & 23 & N/A & N/A & 9 \\
    AP-10K~\cite{yu2021ap}  & 54 & 23 & 10,015 & 17 & N/A & 13,028 & N/A  \\
    Animal track~\cite{zhang2022animaltrack}  & 10 & N/A & 24,700 & N/A & 58 & 429,000 & N/A \\
    APT-36K  & 30 & 15 & 36,000 & 17 & 2,400 & 53,006  & 10\\
    \hline
    \end{tabular}%
  \label{tab:dataset}}%
\end{table}%

\begin{figure}[htbp]
    \centering
    \includegraphics[width=1.0\linewidth]{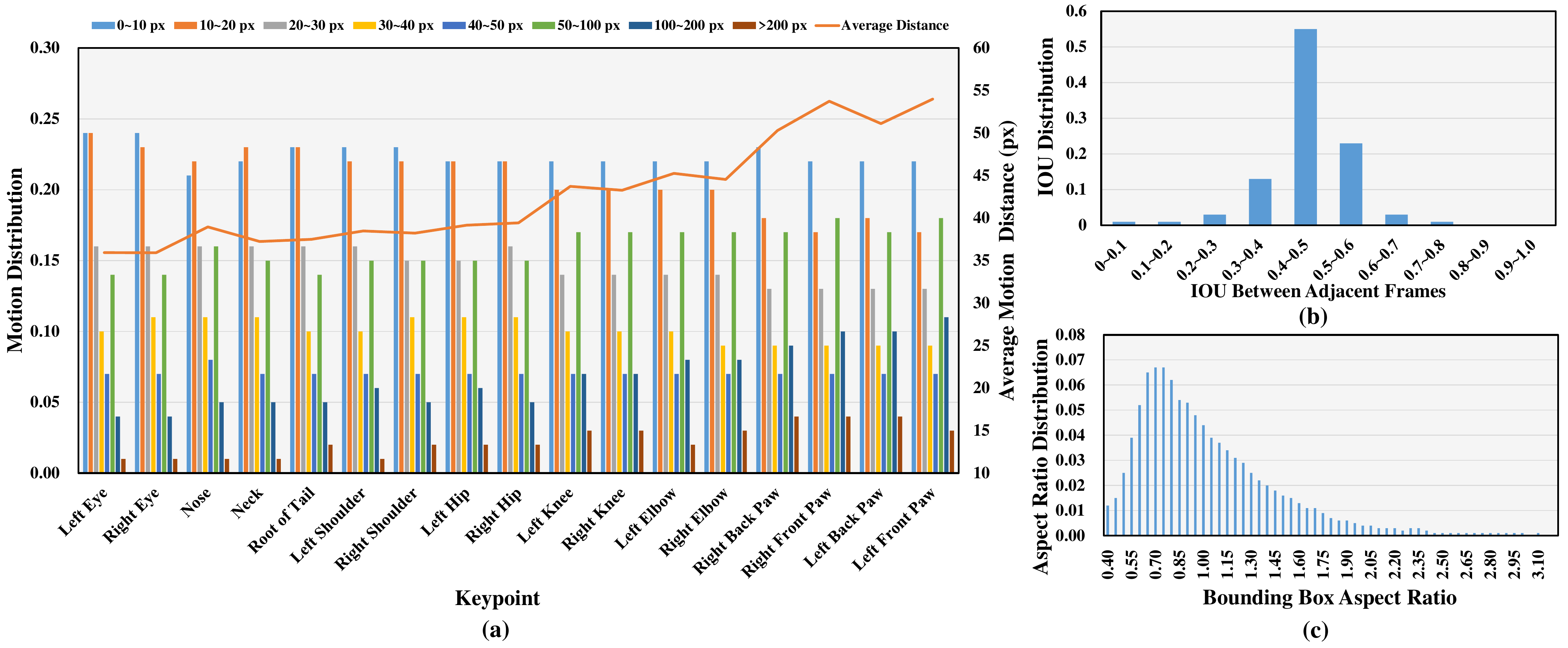}
    \caption{Statistics of APT-36K. (a) The motion distribution and average motion distance of different keypoint categories. "N/A" indicates that the corresponding entry is not available in the given dataset.  (b) The distribution of IOU scores between the tracked bounding boxes in adjacent frames. (c) The distribution of aspect ratios of the bounding boxes in APT-36K.}
    \label{fig:stastic}
\end{figure}

As shown in Table~\ref{tab:dataset}, APT-36K contains 30 different animal species belonging to 15 different families. It has 36,000 annotated frames with 53,006 annotated animal instances from 2,400 video clips, which are much richer than previous animal pose estimation datasets. Therefore, it sets new challenges for animal pose estimation tasks. All the videos are collected from the YouTube website from a set of different topics, including documentary films, vlogs, education films, \etc They are captured using different cameras, at different shooting distances with diverse camera movement patterns. There are 10 types of background in the images of APT-36K, providing diverse scenes for a comprehensive evaluation of animal pose estimation and tracking. For each animal species in APT-36K, there are 80 video clips in total, making it a balanced dataset. It is different from previous animal pose estimation datasets like AP-10K, which is long-tailed and has much fewer instances in some species, \eg, Cercopithecidae. APT-36K is the first dataset suitable for both animal pose estimation and tracking, filling the gap in this area and providing new challenges and opportunities for future research.

We also calculate the distributions of the keypoint motion, IOU between tracked bounding boxes in adjacent frames, and the aspect ratio of the annotated bounding boxes in our APT-36K dataset. As shown in Figure~\ref{fig:stastic} (a), the motion distribution and average motion distance vary a lot for different keypoints, \eg, the average motion distance of paws is over 50 pixels, which is much larger than that of eyes or necks (about 35 pixels). Moreover, the motion magnitudes of shoulder, knee, and hips lie between those of eyes and paws, which is in line with the movement characteristics of four-leg animals. Besides, most of the instances have small IOU scores between their tracked bounding boxes in adjacent frames, implying large motion is very common in APT-36K, as demonstrated in Figure~\ref{fig:stastic} (b). It can also be observed from Figure~\ref{fig:stastic} (c) that the aspect ratio of the bounding box varies a lot from less than 0.4 to more than 3.1. It is because APT-36K contains diverse animals with different actions, \eg, running rabbits and climbing monkeys. These results illustrate the diversity of APT-36K. 

\section{Experiment}
\label{sec:experiment}
\subsection{Implementation details}\label{sec:implementaion}
To provide a comprehensive evaluation for animal pose estimation and tracking, we benchmark representative CNN-based and vision transformer-based pose estimation methods~\cite{sun2019deep,xiao2018simple,xu2022vitpose,yuan2021hrformer} using ground truth bounding boxes or tracked object boxes on the proposed APT-36K dataset. Representative tracking methods~\cite{yan2021learning,li2019siamrpn,lin2021swintrack} are employed to obtain the tracked boxes for the animal instances in the video clips. We set up three tracks based on APT-36K, \ie, the SF track, IS track, and APT track, as described in Sec.~\ref{sec:introduction}. The models are implemented based on the MMPose~\cite{mmpose2020} codebase and trained for 210 epochs following the common practice in human/animal pose estimation tasks. The initial learning rate is 5e-4 and decreased by a factor of 10 at the 170th and 200th epochs. The detailed experimental settings for each track are presented in the following part. We use the average precision (AP)~\cite{lin2014microsoft} as the primary metric to evaluate the performance of different models.
It can be calculated as $AP = \frac{\sum_p \theta(oks_p > T)}{\sum_p 1}$, where $p$ is the index of the person and $oks_p$ is the object keypoint similarity metric. The OKS metric is defined as $ \sum_i[exp(-d_i^2/2s^2k_i^2)\theta(v_i>0)] / \sum_i[\theta(v_i>0)]$, where $d_i$ is the distance between the $i$-th predicted results and the $i$-th ground truth keypoint locations. $s$ is the scale of the object and $k_i$ is a predefined constant that controls falloff. $v_i$ indicates the visibility of the $i$-th keypoint. We use the constant as defined in AP-10K~\cite{yu2021ap}. 

\subsection{Single-frame animal pose estimation (SF track)}
\textbf{Setting} In the SF track, we benchmark the representative CNN-based and vision transformer-based pose estimation methods, including SimpleBaseline~\cite{xiao2018simple}, HRNet~\cite{sun2019deep}, HRFormer~\cite{yuan2021hrformer}, and ViTPose~\cite{xu2022vitpose}. The SimpleBaseline takes the ResNet~\cite{he2016deep} (\ie, ResNet-50 and ResNet-101) as the backbone encoder for feature extraction and uses three deconvolution blocks to up-sample the feature maps for decoding. HRNet takes a similar pipeline but employs a multi-resolution parallel backbone network to extract high-resolution feature maps and discards the deconvolution blocks in the decoder part. HRFormer follows a similar spirit to HRNet and takes a multi-stage transformer structure with multiple branches to jointly encode the multi-resolution information into a high-resolution feature map. ViTPose, on the other hand, utilizes a plain and non-hierarchical vision transformer as a backbone encoder for feature extraction. We set up three settings to benchmark their performance, including using network weights pre-trained on the ImageNet-1K dataset~\cite{deng2009imagenet}, the MS COCO human pose estimation dataset~\cite{lin2014microsoft}, and the AP-10K dataset~\cite{yu2021ap}, respectively. We randomly split the dataset three times with random seeds 0, 10,000, and 20,000, respectively, and train each model accordingly to estimate the error bar of their performance. Specifically, the three settings are detailed as below:
\begin{itemize}
    \item \textit{ImageNet-1K pre-training.} It is a common practice to use ImageNet-1K pre-trained weights to initialize the backbones of pose estimation models. We follow this practice and fine-tune the models initialized with ImageNet-1K pre-trained weights for further 210 epochs on the APT-36K training set. It is noted that we adopt the fully supervised learning scheme on ImageNet-1K to get the pre-trained weights for the backbones used by SimpleBaseline, HRNet, and HRFormer. The vision transformer backbones in ViTPose are initialized with the pre-trained weights from the self-supervised MAE pre-training~\cite{he2021masked}. The Adam~\cite{kingma2014adam} optimizer is utilized for training the CNN-based models, and AdamW~\cite{reddi2018convergence} optimizer is employed to train the vision transformer-based ones, following their default settings.
    \item \textit{Human pose pre-training.} Since the keypoint definition of four-foot animals is similar to those of human beings, it may be beneficial to transfer the knowledge learned from human keypoint annotations to animal pose estimation. Consequently, it will help us make use of the existing large-scale datasets for human pose estimation and bypass the difficulties of building large animal pose estimation datasets. To this end, we pre-train the CNN-based and vision transformer-based models on the MS COCO human pose estimation dataset for 210 epochs. Then, the pre-trained weights are used to initialize the models, which are further fine-tuned on the APT-36K training set for another 210 epochs, following the above ImageNet-1K pre-training setting.
    \item \textit{Animal pose pre-training.} Previous animal pose estimation datasets provide abundant animal images with keypoint annotations for four-foot animals. To investigate the benefit of leveraging the animal pose estimation datasets, we first pre-train the models on AP-10K for 210 epochs and further fine-tune them on the APT-36K training set for another 210 epochs.
\end{itemize}

\begin{table}[htbp]
  \centering
  \scriptsize
  \caption{Results on the APT-36K val set (AP) of different models on the SF track with ImageNet-1K (IN1K)~\cite{deng2009imagenet}, MS COCO~\cite{lin2014microsoft}, and AP-10K~\cite{yu2021ap} pre-training, respectively. $\Delta_{\rm COCO}$ and $\Delta_{\rm AP-10k}$ denote the gains of MS COCO and AP-10K pre-training over ImageNet-1K pre-training, respectively.}
    \setlength{\tabcolsep}{0.01\linewidth}{\begin{tabular}{c|ccccccc}
    \hline
          & {SimpleBaseline} & {SimpleBaseline} & {HRNet} & {HRNet} & {HRFormer} & {HRFormer} & {ViTPose} \\
          & (ResNet-50) & (ResNet-101) & (HRNet-w32) & (HRNet-w48) & (HRFormer-S) & (HRFormer-B) & (ViT-B) \\
    \hline
    IN1K  & 69.4$_{\pm {\rm 1.2}}$  & 69.6$_{\pm {\rm 1.3}}$  & 74.2$_{\pm {\rm 1.1}}$  & 74.1$_{\pm {\rm 0.8}}$  & 71.3$_{\pm {\rm 0.8}}$  & 74.2$_{\pm {\rm 0.9}}$  & 77.4$_{\pm {\rm 1.0}}$ \\
    COCO  & 73.7$_{\pm {\rm 1.2}}$  & 73.5$_{\pm {\rm 1.1}}$  & 76.4$_{\pm {\rm 0.5}}$  & 77.4$_{\pm {\rm 0.7}}$  & 74.6$_{\pm {\rm 1.0}}$  & 76.6$_{\pm {\rm 0.9}}$  & 78.3$_{\pm {\rm 0.8}}$ \\
    $\Delta_{\rm COCO}$ & 4.3   & 3.9   & 2.2   & 3.3   & 3.3   & 2.4   & 0.9 \\
    AP-10K & 72.4$_{\pm {\rm 0.9}}$  & 72.4$_{\pm {\rm 1.0}}$  & 75.9$_{\pm {\rm 1.2}}$  & 76.4$_{\pm {\rm 0.7}}$  & 72.6$_{\pm {\rm 0.9}}$  & 75.2$_{\pm {\rm 0.7}}$  & 78.2$_{\pm {\rm 0.7}}$ \\
    $\Delta_{\rm AP-10K}$ & 3.0     & 2.8   & 1.7   & 2.3   & 1.3   & 1.0     & 0.8 \\
    \hline
    \end{tabular}%
  \label{tab:SF}}%
\end{table}%

\textbf{Results and analysis} The results are summarized in Table~\ref{tab:SF}. It can be observed that with human pose pre-training, both CNN-based and vision transformer-based methods show performance gains, \eg, from 69.6 AP to 73.5 AP for SimpleBaseline with a ResNet-101 backbone network, from 74.1 AP to 77.4 AP for HRNet-w48, and from 74.2 AP to 76.6 AP for HRFormer-B.
A similar benefit can also be obtained by using AP-10K for pre-training, \eg, SimpleBaseline reaches 72.4 AP with either a ResNet-101 or ResNet-50 backbone network, and HRNet-w48 gets a gain of 2.3 AP compared with the ImageNet-1K pre-training. Generally, the benefit is slightly more significant for models with worse performance than those stronger models, which is reasonable.
It is noteworthy that ViTPose with a plain vision transformer backbone, which is pre-trained on ImageNet-1K without using its labels, obtains a remarkable performance of 77.4 AP. Also, after human pose pre-training or animal pose pre-training, the performance could be further improved to 78.3 AP and 78.2 AP, respectively, which is much better than other models.

Another interesting finding is that, although using the animal pose dataset AP-10K for pre-training brings performance gains, the benefit is less than that of using the human pose dataset for pre-training, no matter for CNN-based models or vision transformer-based models (See the third and fifth rows in Table~\ref{tab:SF} denoted by $\Delta_{\rm COCO}$ and $\Delta_{\rm AP-10K}$, respectively). We suspect there are two reasons. First, there is still a domain gap between AP-10K and APT-36K due to their different data sources and distributions, \ie, imbalanced and long-tailed species in AP-10K v.s. balanced ones in APT-36K. Second, MS COCO is much larger than AP-10K by about an order of magnitude, probably leading to more sufficient pre-training since the model could see more diverse training instances and learn more discriminative feature representations. Nevertheless, the difference between $\Delta_{\rm COCO}$ and $\Delta_{\rm AP-10k}$ is not evident for ViTPose, owing to the strong representation ability of vision transformers.

\subsection{Inter-species animal pose generalization (IS track)}

\textbf{Setting} Regarding the diverse animal species in the real world, it is essential to evaluate the inter-species generalization ability of animal pose estimation models, \ie, investigating their performance on unseen animal species. To this end, we set up the IS track, where we select six representative animal families for training and test, \ie, Canidae, Felidae, Hominidae, Cercopithecidae, Ursidae, and Bovidae. In each experiment, all the instances from a specific family form the test set while those instances belong to other families are split into the training set and validation set at a ratio of 9:1. We use the representative HRNet-w32 model with MS COCO pre-training in this track due to its good performance and simplicity. The models are trained following the same setting in the SF track. 

\begin{table}[htbp]
  \centering
  \scriptsize
  \caption{Results of HRNet-w32 models on the IS track (AP) of APT-36K.}
    \setlength{\tabcolsep}{0.02\linewidth}{\begin{tabular}{c|ccccccc}
    \hline
     \diagbox{training}{test} & {Canidae} & {Felidae} & {Hominidae} & {Cercopithecidae} & {Ursidae} & {Bovidae} & {Equidae} \\
    \hline
    \textit{w/o} Canidae  & \uline{\textit{57.6}} & {82.8} & {81.9} & {76.0} &{80.1} & {81.6} & {84.0} \\
    \textit{w/o} Felidae  & {80.9} & \uline{\textit{57.6}} & {81.0} & {76.6} &{79.7} & {81.6} & {84.6} \\
    \textit{w/o} Hominidae  & {80.3} & {81.9} & \uline{\textit{44.2}} & {77.2} &{80.3} & {81.9} & {84.3} \\
    \textit{w/o} Cercopithecidae  & {81.1} & {83.4} & {80.8} & \uline{\textit{29.6}} &{80.7} & {81.1} & {84.7} \\
    \textit{w/o} Ursidae  & {80.7} & {83.3} & {80.8} & {76.7} &\uline{\textit{43.3}} & {82.0} & {84.1} \\
    \textit{w/o} Bovidae  & {80.1} & {83.6} & {80.6} & {76.7} &{80.8} & \uline{\textit{58.6}} & {84.5} \\
    \textit{w/o} Equidae  & {80.4} & {82.4} & {82.0} & {77.2} &{79.7} &{82.2} & \uline{\textit{61.5}}  \\
    \hline
    \end{tabular}%
  \label{tab:IS}}%
\end{table}%

\textbf{Results and analysis} As can be seen from Table~\ref{tab:IS}, the models generalize well on the Canidae, Felidae, Bovidae, and Equidae families with instances from other animals for training. For example, it obtains 57.6 AP on the Canidae family, as indicated in the top-left cell. It is because although the animal instances from the Canidae family are not used during training, they share some commonness with animals in the Ursidae family and Felidae family since they belong to the same Carnivora order. For rare species that do not share commonness with the training set, the models show much poorer generalization ability, \eg, the model trained without data from the Cercopithecidae family only obtains 29.6 AP on the Cercopithecidae family. In contrast, after using the instances from the Cercopithecidae family for training, the performance could reach over 76.0 AP (See other scores in the fourth column except the diagonal one). It also should be noted that although the model generalizes well on the families mentioned above, the performance still falls behind the models that have seen data from those families during training by a large margin, \eg, 57.6 AP v.s. over 80 AP on the Canidae family. The results imply that 1) each animal species has its own characteristics, 2) it is beneficial and necessary (if possible) to collect and annotate animal instances from diverse animal families, especially for rare species like the Cercopithecidae family, and 3) more efforts should be made to improve the inter-species generalization ability of animal pose estimation models. Besides, the models trained with slightly different training sets demonstrate similar performance on the same seen family as shown in each column (except the diagonal one) in Table~\ref{tab:IS}, which is probably attributed to the balanced data distribution (80 video clips for each species) of animal species in the APT-36K dataset.

\begin{table}[htbp]
  \centering
  \scriptsize
  \caption{Results of HRNet-w32 models on the few-shot learning setting (AP) of APT-36K.}
    \setlength{\tabcolsep}{0.02\linewidth}{\begin{tabular}{c|ccccccc}
    \hline
     {} & {Canidae} & {Felidae} & {Hominidae} & {Cercopithecidae} & {Ursidae} & {Bovidae} & {Equidae} \\
    \hline
    \textit zero-shot & {52.3} & {54.8} & {46.3} &{35.6} & {38.0} & {54.9} &{61.4} \\
    \textit 20-shot & {53.3} & {55.5} & {48.4} &{37.6} & {43.3} & {55.5} &{61.7} \\
    \textit 30-shot & {53.9} & {56.0} & {52.5} &{41.6} & {48.0} & {56.3} &{62.0} \\
    \textit 40-shot & {54.3} & {57.2} & {52.7} &{42.1} & {48.1} & {56.8} &{62.8} \\
    \hline
    \end{tabular}%
  \label{tab:fs}}%
\end{table}%

\textbf{Few-shot learning} To further evaluate the model's generalization ability, we carried out the experiment under the few-shot setting. As can be seen from Table~\ref{tab:fs}, with more data used for training, the performance is greatly improved, especially on species with unique textures, appearance, and posture characteristics, e.g., the Hominidae, Cercopithecidae, and Ursidae species. It is because these unique characteristics are not shared in the training images of other species. The performance gain brought by more training data is relatively smaller for species that shares some common characteristics with the training species, e.g., the Canidae, Felidae, Bovidae, and Equidae species. We think that the few-shot setting is an important research topic in animal pose estimation, e.g., how to make the pre-trained models generalize well on unseen species. Besides, recent studies have shown that large models are already few-shot learners~\cite{zhang2022vitaev2,brown2020language}. It is interesting to explore their performance on the few-shot animal pose estimation tasks, where the proposed dataset can provide a suitable benchmark.

\subsection{Animal pose tracking (APT track)}

\textbf{Setting} In this track, we use representative object trackers with both CNN-based backbones and vision transformer-based backbones to track each animal instance across the video clips, giving each animal's ground truth bounding box in the first frame. Once the tracked bounding boxes are obtained, the pose estimation methods with MS COCO pre-training are used for animal pose estimation accordingly. We also use the average precision metric for evaluation. Specially, the CNN-based tracking methods SiamRPN++~\cite{li2019siamrpn} and STARK~\cite{yan2021learning} with a ResNet-50 backbone~\cite{he2016deep} are employed for animal tracking. For vision transformer-based methods, we adopt the recent SOTA tracking method SwinTrack~\cite{lin2021swintrack} with a Swin transformer backbone~\cite{liu2021swin}. We also design a simpler \textbf{ViTTrack} baseline with the plain vision transformer as the backbone, \ie, ViT~\cite{dosovitskiy2020image}, to compare the performance of both stage-wise transformer structure and plain transformer structure. ViTTrack is a siamese structure with a shared backbone encoder for feature extraction. The target object in the first frame is used as a template for feature matching in the subsequent frames and tracking. Specifically, the features from subsequent frames are concatenated with the template feature and fed into decoder layers, whose output is then used to predict the locations of the target object via a simple MLP. The encoder and decoder are all plain vision transformers, which are pre-trained on ImageNet-1K via MAE~\cite{he2021masked}. To reduce the computational cost, the template image size is usually set to 112$\times$112, while the size of the search region in the subsequent frames is set to 224$\times$224.

\begin{table}[htbp]
  \centering
  \scriptsize
  \caption{Results on the APT-36K test set (AP) of different models on the APT track with different object trackers. $\dag$ denotes ViTTrack uses the fixed ViT encoder of ViTPose trained on APT-36K.}
    \setlength{\tabcolsep}{0.01\linewidth}{\begin{tabular}{c|ccccccc}
    \hline
          & SimpleBaseline & SimpleBaseline & HRNet & HRNet & HRFormer & HRFormer & ViTPose \\
          & (ResNet-50) & (ResNet-101) & (HRNet-w32) & (HRNet-w48) & (HRFormer-S) & (HRFormer-B) & (ViT-B) \\
    \hline
    SiamRPN++~\cite{li2019siamrpn} & 70.2$_{\pm {\rm 1.7}}$  & 70.1$_{\pm {\rm 1.6}}$  & 73.0$_{\pm {\rm 1.4}}$  & 73.6$_{\pm {\rm 1.6}}$  & 70.9$_{\pm {\rm 1.6}}$  & 73.1$_{\pm {\rm 1.4}}$  & 74.2$_{\pm {\rm 1.1}}$  \\
    STARK~\cite{yan2021learning} & 71.5$_{\pm {\rm 1.7}}$  & 71.4$_{\pm {\rm 1.7}}$  & 74.1$_{\pm {\rm 1.4}}$  & 74.8$_{\pm {\rm 1.5}}$  & 72.1$_{\pm {\rm 1.5}}$  & 74.2$_{\pm {\rm 1.4}}$  & 75.3$_{\pm {\rm 1.0}}$  \\
    SwinTrack~\cite{lin2021swintrack} & 71.6$_{\pm {\rm 1.8}}$  & 71.5$_{\pm {\rm 1.6}}$  & 74.1$_{\pm {\rm 1.4}}$  & 74.9$_{\pm {\rm 1.6}}$  & 72.2$_{\pm {\rm 1.8}}$  & 74.3$_{\pm {\rm 1.5}}$  & 75.4$_{\pm {\rm 1.1}}$  \\
    ViTTrack & 71.9$_{\pm {\rm 1.6}}$  & 71.9$_{\pm {\rm 1.4}}$  & 74.4$_{\pm {\rm 1.2}}$  & 75.3$_{\pm {\rm 1.4}}$  & 72.7$_{\pm {\rm 1.4}}$  & 74.6$_{\pm {\rm 1.2}}$  & 75.8$_{\pm {\rm 0.9}}$  \\
    ViTTrack$^\dag$ & 71.7$_{\pm {\rm 1.7}}$  & 71.6$_{\pm {\rm 1.4}}$  & 74.2$_{\pm {\rm 1.1}}$  & 74.9$_{\pm {\rm 1.4}}$  & 72.3$_{\pm {\rm 1.4}}$  & 74.5$_{\pm {\rm 1.2}}$  & 75.5$_{\pm {\rm 0.9}}$  \\
    \hline
    \end{tabular}}%
  \label{tab:track}%
\end{table}%

\textbf{Results and analysis} The results are summarized in Table~\ref{tab:track}. It can be observed that the APT performance based on the vision transformer-based trackers is slightly better compared with that using CNN-based trackers, \ie, HRFormer-S~\cite{yuan2021hrformer} obtains 72.1 AP with STARK~\cite{yan2021learning} and 72.7 AP with ViTTrack, respectively. Similarly, SimpleBaseline~\cite{xiao2018simple} with ResNet-50~\cite{he2016deep} achieves 71.5 AP with STARK and 71.9 AP with ViTTrack. Again, ViTPose achieves the best performance among all the pose estimation models no matter which object tracker is used. Moreover, ViTPose with ViTTrack delivers the best 75.8 AP, surpassing the SimpleBaseline with SiamRPN++ tracker by a large margin of 5.6 AP. Besides, we initialize the ViT-B encoder of ViTTrack with the weights of the ViT-B encoder in ViTPose trained on APT-36K training set and keep it fixed during training on the tracking data, \ie, denoted as ViTTrack$^\dag$ in Table~\ref{tab:track}. Surprisingly, with the shared ViT-B encoder, ViTPose with ViTTrack$^\dag$ obtains the second-best performance, \ie, 75.5 AP, which is even better than ViTPose with SwinTrack, which requires an extra Swin transformer encoder. These results imply the potential of plain vision transformers as a foundation model for simultaneously serving multiple vision tasks, which is of great significance and deserves more research in future work. 

\begin{table}[htbp]
  \centering
  \scriptsize
  \caption{Results of different single object tracking methods (success and precision) on APT-36K.}
    \setlength{\tabcolsep}{0.02\linewidth}{\begin{tabular}{c|ccccccc}
    \hline
     {} & {SwinTrack~\cite{lin2021swintrack}} & {SiamRPN~\cite{li2019siamrpn}} & {STRAK~\cite{yan2021learning}} & {ViTTrack} & {ViTTrack$^\dag$} \\
    \hline
    \textit Success & {81.5} & {76.3} & {81.9} &{81.2} & {80.9} \\
    \textit Precision & {62.9} & {46.8} & {62.6} &{64.2} & {61.5}  \\
    \hline
    \end{tabular}%
  \label{tab:mt}}%
\end{table}%

Multi-object tracking methods have been studied in previous pose-tracking tasks. However, it is really a hard job to adapt existing multi-object tracking methods to the animal pose tracking task, due to the scope mismatching issue between the animal species and in-distribution categories of current detectors. Specifically, most of the animals in the proposed APT-36K dataset can't be successfully recognized by the current object detectors. To address this issue, a simplified strategy in SimpleBaseline~\cite{xiao2018simple} using optical flow-based tracking is adopted to evaluate the performance of multi-object tracking on APT-36K. With the HRNet-w32~\cite{sun2019deep} pose estimator, this strategy obtains 71.3 AP, which is not well as the single-object tracking results, \eg, at least 73.0 AP as shown in the 3rd column in Table~\ref{tab:track}. Thus, we focus on single-object tracking in this paper and compare their success rate and precision using the OPE metric~\cite{wu2013online}. The results are summarized in Table~\ref{tab:mt}. It can be observed that the success rate of single object tracking on the APT-36K dataset is around 80 and the precision is slightly above 60 even for the strong tracker SwinTrack~\cite{lin2021swintrack}. Such results demonstrate that the proposed APT-36K dataset is challenging for both multi-object and single-object tracking methods. How to efficiently deal with the diverse animal species deserves more research efforts in the future, \eg, an object detector with better detection results for out-of-distribution categories. 

\section{Limitation and discussion}
\label{sec:lim}
Our APT-36K fills the gap between single-frame animal pose estimation and animal tracking datasets, based on which we benchmarked representative pose estimation models and gained some insights. We believe APT-36K can benefit the future research of animal pose estimation and tracking, \eg, regarding novel model design, effective inter-species generalization, and multi-task learning. Although the scale of APT-36K is larger than previous animal pose estimation datasets, it is still much smaller than human pose estimation datasets. Considering the diverse species of animals, there are more efforts to be made in future work. Besides, the number of average animal instances in the video clips of APT-36K is limited compared with that in animal instance tracking datasets. Although more animals in the video clips always imply many small and occluded instances, it makes the annotation process of keypoint tracking extremely more difficult and expensive compared with annotating the bounding boxes of animal instances. To this end, it still matters to train better animal pose tracking models to be applicable in real-world scenarios. One possible solution is to resort to a modern 3D engine to generate realistic synthetic images and automatically produce annotations automatically.

\section{Conclusion}

We establish APT-36K, \ie, the first large-scale dataset with high-quality animal pose annotations from consecutive frames for animal pose estimation and tracking. Based on APT-36K, we benchmark representative state-of-the-art pose estimation methods under the single-frame animal pose estimation setting, inter-species animal pose generalization setting, and animal pose tracking setting, respectively. Extensive experimental results demonstrate the benefit of inter- and intra-domain pre-training for animal pose estimation, the significance of collecting and annotating keypoints of diverse animal species, and the great potential of plain vision transformers for animal pose tracking. We hope APT-36K can provide new opportunities for further animal pose estimation and tracking research.

\textbf{Social impact.} The proposed APT-36K can facilitate the research of animal pose estimation and tracking, which is beneficial for animal behavior understanding and wildlife preservation. However, due to the great diversity of real-world animal species, it should be careful about whether a model trained on APT-36K can generalize well on unseen rare animal species.

\acksection
The project is funded by the National Natural Science Foundation of China under Grant 61873077.

\appendix

\section{Subjective results}

\begin{figure}[htbp]
    \centering
    \includegraphics[width=\linewidth]{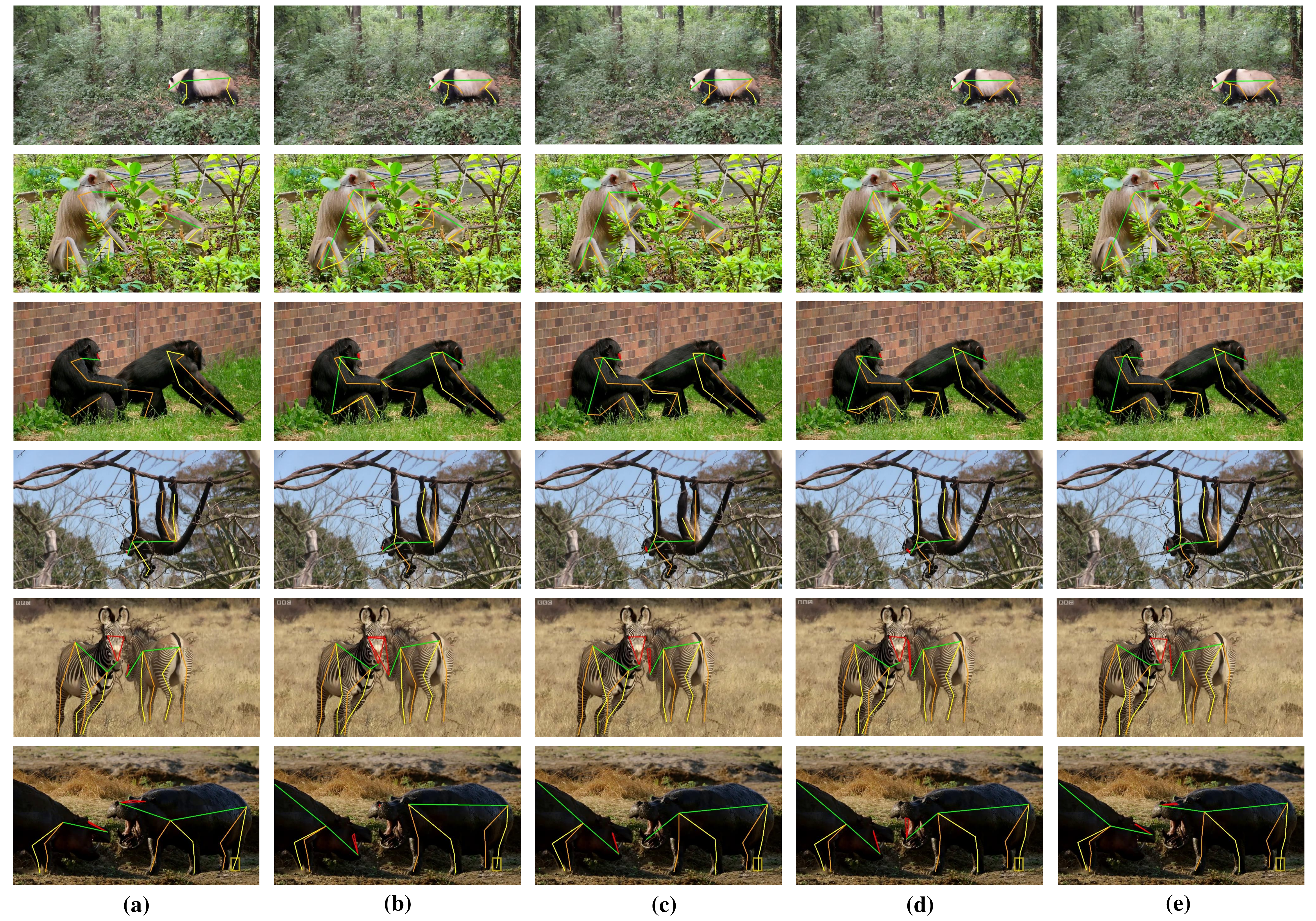}
    \caption{Subjective results of (a) groundtruth; (b) SimpleBaseline~\cite{xiao2018simple} with ResNet-101~\cite{he2016deep}; (c) HRNet-w48~\cite{sun2019deep}; (d) HRFormer-B~\cite{yuan2021hrformer}; and (e) ViTPose~\cite{xu2022vitpose}. The video-based results from ViTPose are provided in the supplementary video.}
    \label{fig:subjective}
\end{figure}

To subjectively evaluate the results of different pose estimation methods trained on the APT-36K dataset, we visualize the results generated by the representative methods SimpleBaseline~\cite{xiao2018simple} with ResNet-101~\cite{he2016deep} backbone, HRNet-w48~\cite{sun2019deep}, HRFormer-B~\cite{yuan2021hrformer}, and ViTPose~\cite{xu2022vitpose} with ViT-B~\cite{dosovitskiy2020image} backbone. All models are pre-trained with the human pose estimation data from the MS COCO~\cite{lin2014microsoft} dataset. As demonstrated in Figure~\ref{fig:subjective}, the methods trained with the APT-36K dataset successfully predict the keypoints of different animal species, despite challenging cases like occlusion (the 2nd and 3rd row). They can also deal with animals with irregular body postures like in the 4th row and different orientations like in the 5th row. 
It can also be observed that ViTPose can better deal with challenging cases like multiple instances with irregular body postures. As shown in the last row of Figure~\ref{fig:subjective}, ViTPose predicts more precise eye and nose locations for the hippopotamus at the right.

\begin{table}[htbp]
  \centering
  \scriptsize
  \caption{Detail animal species and classification information in APT-36K.}
    \setlength{\tabcolsep}{0.02\linewidth}{\begin{tabular}{cccccccc}
    \hline
      {Felidae} & {Bovidae} & {Canidae} & {Hominidae} & {Cercopithecidae} & {Ursidae}  & {Equidae}  & {Cervidae}  \\
    \hline
     {cat} & {antelope} & {dog} &{chimpanzee} & {monkey} & {panda} &{horse} &{deer}\\
     {tiger} & {buffalo} & {fox} &{orangutan} & {spider-monkey} & {black-bear} &{zebra} \\
     {cheetah} & {cow} & {wolf} &{gorilla} & {howling-monkey} & {polar-bear} &{} \\
     {lion} & {sheep} & {} &{} & {} & {} &{} \\
    \hline
     {} & {} & {} &{} & {} & {} &{} \\
    \hline
    {Leporidae} & {Suidae} & {Elephantidae} & {Hippopotamidae} & {Procyonidae} & {Rhinocerotidae}  & {Giraffidae} \\
    \hline
     {rabbit} & {pig} & {elephant} &{hippo} & {raccoon} & {rhino} &{giraffe} \\
    \hline
    \end{tabular}%
  \label{tab:if}}%
\end{table}%

\section{Video clip choosing}

To guarantee the diversity and annotation quality of the proposed APT-36K dataset, we select 30 different animal species with distinct features that people are familiar with to label. Animals that do not satisfy these requirements, such as hamsters and beavers, are not considered since these animals are indistinguishable when moving. To ensure a reasonable difficulty distribution of the dataset, we follows the 5:3:2 principles to select the videos, \ie, 50$\%$ of the selected videos are the simple cases which contain a single animal without occlusion; 30$\%$ of them can be categorized as the medium-difficult cases which involves a single animal with occlusion; and the left 20$\%$ of them are difficult cases with multiple animals and occlusion. Meanwhile, to ensure the diversity of the backgrounds, we strictly control that there are fewer than two clips having the same kind of background.

\section{Annotators training}

To train the annotators, we first employ animal skeletal models with annotation examples to teach the annotators how to annotate the keypoints for each animal. After that, we assign each annotator several test video clips for annotation. The annotation results on these test clips will be compared with the ground truth keypoint annotations to select the appropriate annotators in the following annotation process. During the annotation, we first have the annotators annotate each clip. The annotation results are checked by the senior annotators. After that, the checked results are further validated by the organizer to further reduce the errors in the annotations. To deal with challenging cases like occlusion, it should be noted that there are adjacent frames for a given frame to be annotated in a video. In this case, annotators can estimate the occluded keypoints via the clues from several adjacent frames, where the keypoints are visible and not-occluded. In addition, for the keypoints that are heavily occluded and can not be determined by the annotators with high confidence, we mark them as obscured and do not take them into account for model training and evaluation.

\section{Memory and inference speed comparison}

\begin{table}[htbp]
  \centering
  \scriptsize
  \caption{Training memory and inference speed comparison of SimpleBaseline~\cite{xiao2018simple}, HRNet~\cite{sun2019deep}, HRFormer~\cite{yuan2021hrformer}, and ViTPose~\cite{xu2022vitpose} on the APT-36K dataset.}
    \setlength{\tabcolsep}{0.003\linewidth}{\begin{tabular}{c|ccccccc}
    \hline
          & ResNet-50~\cite{xiao2018simple} & ResNet-101~\cite{xiao2018simple} & HRNet-w32~\cite{sun2019deep} & HRNet-w48~\cite{sun2019deep} & HRFormer-S~\cite{yuan2021hrformer} & HRFormer-B~\cite{yuan2021hrformer} & ViTPose-B~\cite{xu2022vitpose} \\
    \hline
    Training Memory (M) & 10,041 & 13,589 & 15,051 & 19,635 & 28,888 & OOM   & 19,715 \\
    Throughput (FPS) & 1,102 & 834   & 719   & 521   & 208   & 123   & 711 \\
    Precision (AP) & 73.7  & 73.5  & 76.4  & 77.4  & 74.6  & 76.6  & 78.3 \\
    \hline
    \end{tabular}}%
  \label{tab:memory}%
\end{table}%

We also compare these pose estimation methods' training memory footprint and inference speed on A100 machines with 40G memory. We use a batch size of 64 input images with a resolution of 256*256 during both training and inference. 
The results are summarized in Table~\ref{tab:memory}. It can be observed that the HRFormer~\cite{yuan2021hrformer} variants consume much more memory compared with other methods due to the quadratic memory consumption of transformer layers and the high-resolution feature map employed in their structures. The CNN-based framework SimpleBaseline~\cite{xiao2018simple} favors the inference speed and performs a little worse on the pose estimation accuracy than other methods. ViTPose obtains a better trade-off between the inference speed and accuracy, \ie, it obtains 78.3 AP with 711 frame-per-second (FPS), while HRNet-w48~\cite{sun2019deep} obtains 77.4 AP with a lower 512 FPS.

\section{More implementation details}

\begin{figure}
    \centering
    \includegraphics[width=\linewidth]{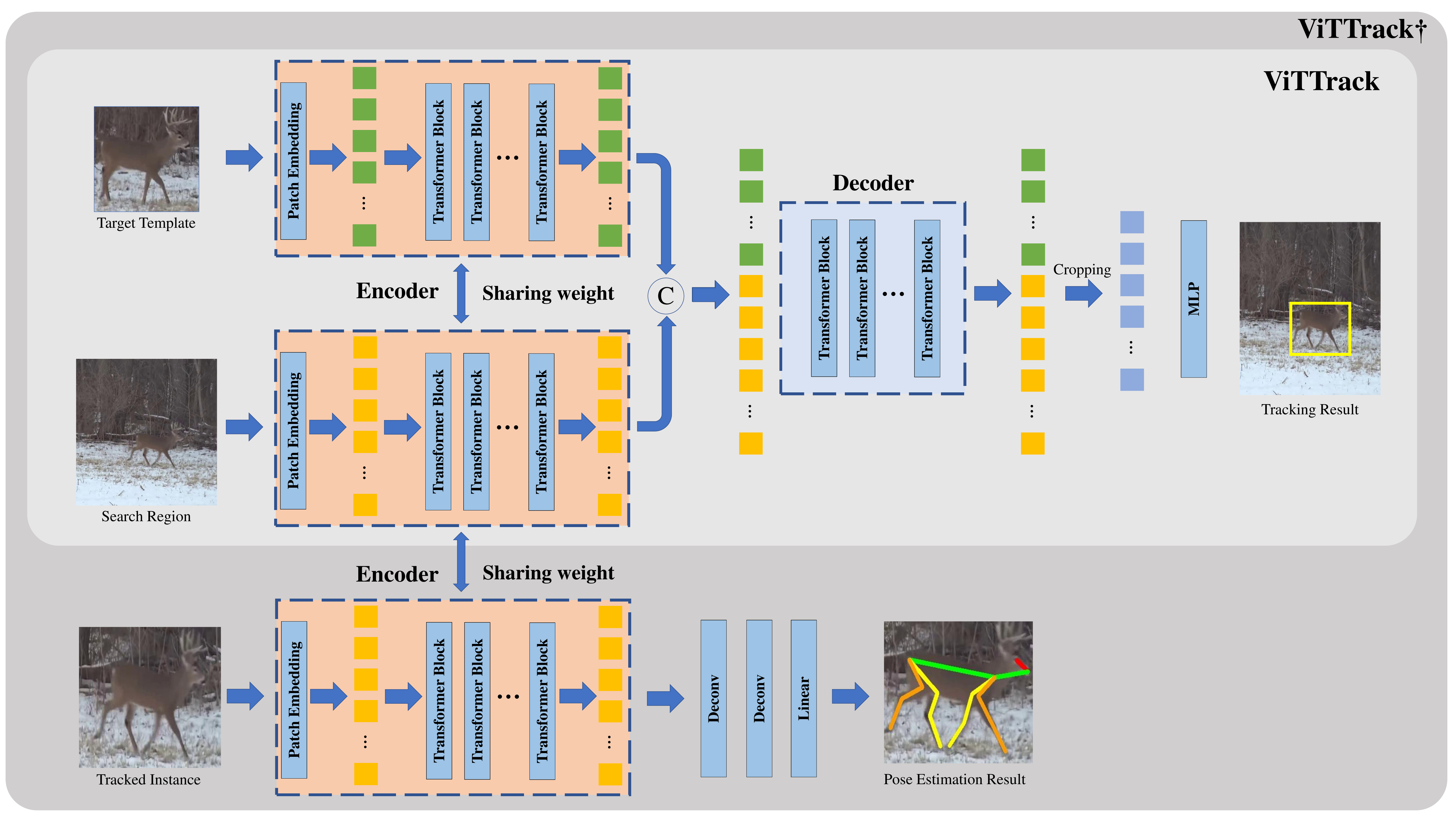}
    \caption{The structures of the proposed ViTTrack and ViTTrack$^\dag$. We employ a plain vision transformer based encoder-decoder structure for object tracking in ViTTrack. To explore the potential of vision transformers for multi-task learning, we design a simple ViTTrack$^\dag$ model by using a shared backbone for pose estimation and tracking.}
    \label{fig:vittrack}
\end{figure}

We train all the pose estimation models on 8 Nvidia A100 GPUs for 210 epochs, with the implementation details provided in the main text. We also provide a detailed description of the proposed ViTTrack framework in this section. As demonstrated in Figure~\ref{fig:vittrack}, ViTTrack takes a siamese structure to extract features from the template and the search region with a weight-sharing encoder. The encoder is a plain vision transformer structure initialized with MAE pretraining on the ImageNet-1K training set. Then, the template and search region features are concatenated and fed into the decoder, which is also composed of several transformer layers with self-attention. To fully utilize the power of pretraining, we adopt exactly the same structure for the decoder from the MAE pretraining and initialize it accordingly. It is a reasonable design as the target of the decoder is modeling the similarity between input tokens no matter in the upstream masked image pretraining tasks and downstream visual object tracking task, \ie, modeling the similarity between the template and the search regions. After that, the tokens corresponding to the search regions are selected and used to predict the location of the tracked instance. The ViTTrack model has been trained for 300 epochs with the GOT-10K~\cite{huang2019got} dataset, following the same setting in SwinTrack~\cite{lin2021swintrack}. For the ViTTrack$^\dag$ framework, we first train the ViTPose model on the APT-36K dataset for 210 epochs with the animal pose estimation data. Then, the backbone weights of ViTPose are utilized to initialize the encoder in ViTTrack as they are the same ViT-B structure. After that, we train the ViTTrack's decoder with the GOT-10K dataset for 300 epochs, with the encoder part frozen. Thus, the pose estimator and tracker share the same encoder for feature extraction and separately use task-specific decoders for pose estimation and tracking. We denote the multi-task framework as ViTTrack$^\dag$.

\section{Datasheet}
\subsection{Motivation}

\noindent \textbf{1. For what purpose was the dataset created? Was there a specific task in mind? Was there a specific gap that needed to be filled? Please provide a description.}

\textbf{A1:} 
APT-36K is created to facilitate the development and evaluation of video-based animal pose estimation and tracking methods. Previous datasets either focused on animal tracking or single-frame-based animal pose estimation, and neither did both. Based on the proposed APT-36K dataset, we try to answer several challenging questions: 1) whether previous human or animal pose datasets benefit animal pose tracking tasks; 2) How well do animal pose estimation methods generalize across species in a balanced animal distribution setting; and 3) whether the animal pose estimation methods work well in animal pose tracking setting. The answer to the animal pose tracking task is impossible to explore with previous datasets as they do not consider the animal pose tracking scenarios. Besides, APT-36K is the first animal pose estimation dataset with balanced animal species distribution, with abundant species corresponding to long-tailed species in previous datasets~\cite{yu2021ap}. Thus, we believe APT-36K can contribute to the development of animal pose tracking algorithms and single-frame-based animal pose estimation.

\noindent \textbf{2. Who created this dataset (e.g., which team, research group) and on behalf of which entity (e.g., company, institution, organization)?}

\textbf{A2:} APT-36K is created by the authors as well as some volunteer graduate students from Hangzhou Dianzi University, including Jiahao Jiang, Xuepu Zeng, Jiamian Xu, Jiacheng Zhang, Yuhu Xin, Jiajie Huang, Jingsheng Fang.

\noindent \textbf{3. Who funded the creation of the dataset? If there is an associated grant, please provide the name of the grantor and the grant name and number.}

\textbf{A3:} The project is funded by the National Natural Science Foundation of China under Grant 61873077.

\subsection{Composition}

\noindent \textbf{1. What do the instances that comprise the dataset represent (e.g., documents, photos, people, countries)? Are there multiple types of instances (e.g., movies, users, and ratings; people and interactions between them; nodes and edges)? Please provide a description.}

\textbf{A1:} APT-36K consists of 2,400 video clips collected and filtered from 30 animal species with 15 frames for each video, resulting in 36,000 frames in total. The animal species are organized under taxonomic ranks. For each animal instance in the videos, we annotate its location, tracking identifier, and 17 keypoints (left eye, right eye, nose, neck, tail root, left shoulder, left elbow, left front paw, right shoulder, right elbow, right front paw, left hip, left knee, left hind paw, right hip, right knee, right hind paw). We also annotate the type of the background in the frames. The annotations are organized in the MS COCO~\cite{lin2014microsoft} format.

\noindent \textbf{2. How many instances are there in total (of each type, if appropriate)?}

\textbf{A2:} The APT-36K dataset contains 36,000 images and 53,006 instances with high-quality position, tracking identifier, and keypoint annotations. 

\noindent \textbf{3. Does the dataset contain all possible instances or is it a sample (not necessarily random) of instances from a larger set? If the dataset is a sample, then what is the larger set? Is the sample representative of the larger set (e.g., geographic coverage)? If so, please describe how this representativeness was validated/verified. If it is not representative of the larger set, please describe why not (e.g., to cover a more diverse range of instances, because instances were withheld or unavailable).}

\textbf{A3:} APT-36K collects and filters the videos from the YouTube website and covers 30 animal species selected from the real-world animal species. Due to the high diversity and huge amounts of animal species in the real world, it is almost impossible to include all animal species in a single dataset. We will continue to increase the diversity and volume of APT-36K in future work.

\noindent \textbf{4. What data does each instance consist of? “Raw” data (e.g., unprocessed text or images)or features? In either case, please provide a description.}

\textbf{A4:} Each instance consists of one animal with its RGB descriptions and annotations, including location (bounding box) annotation, family and species annotations, keypoints annotations, background type and tracking identifier annotations.

\noindent \textbf{5. Is there a label or target associated with each instance? If so, please provide a description.}

\textbf{A5:} Yes. Each instance is labeled with instance ID, image ID, animal species information (family and species), area of location box, keypoint information, number of keypoints, background category, the source of images and videos, and its unique tracking ID in the video, according to COCO labeling style.

\noindent \textbf{6. Is any information missing from individual instances? If so, please provide a description, explaining why this information is missing (e.g., because it was unavailable). This does not include intentionally removed information, but might include, e.g., redacted text.}

\textbf{A6:} Yes. We omit the possible low-quality annotations for some instances' keypoints, which may be heavily obscured, blurred, or small in scale. It is a common practice to improve the annotation quality as described in establishing the COCO human pose dataset. Following the labeling routine in MS COCO, we mark the confidence of these keypoints as zero in the annotation files. 

\noindent \textbf{7. Are relationships between individual instances made explicit (e.g., users’ movie ratings, social network links)? If so, please describe how these relationships are made explicit.}

\textbf{A7:} Yes. The annotations are labeled following the MS COCO labeling routine. Thus, the relationship between each instance, \eg, whether two instances are located on the same image, can be queried using the \href{https://github.com/jin-s13/xtcocoapi}{COCO APIs}.

\noindent \textbf{8. Are there recommended data splits (e.g., training, development/validation, testing)? If so, please provide a description of these splits, explaining the rationale behind them.}

\textbf{A8:} Yes. The dataset is split into three disjoint subsets for training, validation, and test, respectively, following the ratio of 7:1:2 per animal species. It is also noteworthy that we adopt a video-level partition to prevent the potential information leakage since the frames in the same video clip are similar to each other.

\noindent \textbf{9. Are there any errors, sources of noise, or redundancies in the dataset? If so, please provide a description.}

\textbf{A9:} Although we have double-checked the annotation information very carefully, there may be some inaccurate keypoint annotations, such as minor drift in the keypoint positions, due to the occlusion or blur caused by animals' moving.

\noindent \textbf{10. Is the dataset self-contained, or does it link to or otherwise rely on external resources (e.g., websites, tweets, other datasets)? If it links to or relies on external resources, a) are there guarantees that they will exist, and remain constant, over time; b) are there official archival versions of the complete dataset (i.e., including the external resources as they existed at the time the dataset was created); c) are there any restrictions (e.g., licenses, fees) associated with any of the external resources that might apply to a future user? Please provide descriptions of all external resources and any restrictions associated with them, as well as links or other access points, as appropriate.}

\textbf{A10:} Yes. APT-36K does not use content from other existing datasets. The video clips used in APT-36K are from publicly available videos on the Youtube dataset. We acknowledge the video creators' efforts and attach the links to the source videos in the APT-36K dataset.

\noindent \textbf{11. Does the dataset contain data that might be considered confidential (e.g., data that is protected by legal privilege or by doctorpatient confidentiality, data that includes the content of individuals non-public communications)? If so, please provide a description.}

\textbf{A11:} No.

\noindent \textbf{12. Does the dataset contain data that, if viewed directly, might be offensive, insulting, threatening, or might otherwise cause anxiety? If so, please describe why.}

\textbf{A12:} No.

\subsection{Collection Process}

\noindent \textbf{1. How was the data associated with each instance acquired? Was the data directly observable (e.g., raw text, movie ratings), reported by subjects (e.g., survey responses), or indirectly inferred/derived from other data (e.g., part-of-speech tags, model-based guesses for age or language)? If data was reported by subjects or indirectly inferred/derived from other data, was the data validated/verified? If so, please describe how.}

\textbf{A1:} The keypoint information is observable for each animal. Well-trained annotators are recruited to label the locations of the animal keypoints directly with annotation tools like \href{https://github.com/wkentaro/labelme}{labelme}.

\noindent \textbf{2. What mechanisms or procedures were used to collect the data (e.g., hardware apparatus or sensor, manual human curation, software program, software API)? How were these mechanisms or procedures validated?}

\textbf{A2:} The video clips in APT-36K are all selected and filtered from publicly available videos on Youtube. We manually select the representative clips from these videos, where each clip contains observable animal movement.

\noindent \textbf{3. If the dataset is a sample from a larger set, what was the sampling strategy (e.g., deterministic, probabilistic with specific sampling probabilities)?}

\textbf{A3:} No.

\noindent \textbf{4. Who was involved in the data collection process (e.g., students, crowdworkers, contractors) and how were they compensated (e.g., how much were crowdworkers paid)?}

\textbf{A4:} The authors of the paper.

\noindent \textbf{5. Over what timeframe was the data collected? Does this timeframe match the creation timeframe of the data associated with the instances (e.g., recent crawl of old news articles)? If not, please describe the timeframe in which the data associated with the instances was created.}

\textbf{A5}: It took about 15 days to collect the data and about 3 months to complete the data cleaning, organization, and annotation process.

\subsection{Preprocessing/cleaning/labeling}

\noindent \textbf{1. Was any preprocessing/cleaning/labeling of the data done (e.g., discretization or bucketing, tokenization, part-of-speech tagging, SIFT feature extraction, removal of instances, processing of missing values)? If so, please provide a description. If not, you may skip the remainder of the questions in this section.}

\textbf{A1:} We manually collect and clean the data from the YouTube website to ensure there are high-quality images in the APT-36K dataset with 1920*1080 resolution. Besides, we have each annotator manually check videos containing several animal instances to remove the duplicate videos in the dataset.

\noindent \textbf{2. Was the ``raw" data saved in addition to the preprocessed/cleaned/labeled data (e.g., to support unanticipated future uses)? If so, please provide a link or other access point to the ``raw" data.}

\textbf{A2:} No. We provide the links to the source videos in the annotation files.

\noindent \textbf{3. Is the software used to preprocess/clean/label the instances available? If so, please provide a link or other access point.}

\textbf{A3:} We use the open source labeling tool \href{https://github.com/wkentaro/labelme}{labelme}.

\subsection{Uses}

\noindent \textbf{1. Has the dataset been used for any tasks already? If so, please provide a description.}

\textbf{A1:} No.

\noindent \textbf{2. Is there a repository that links to any or all papers or systems that use the dataset? If so, please provide a link or other access point.}

\textbf{A2:} N/A.

\noindent \textbf{3. What (other) tasks could the dataset be used for?}

\textbf{A3:} APT-36K can be used for both animal pose estimation and animal pose tracking studies. In addition, it can be used for topics such as animal species classification as well as animal behavior understanding and behavior prediction (with further corresponding annotations). 

\noindent \textbf{4. Is there anything about the composition of the dataset or the way it was collected and preprocessed/cleaned/labeled that might impact future uses? For example, is there anything that a future user might need to know to avoid uses that could result in unfair treatment of individuals or groups (e.g., stereotyping, quality of service issues) or other undesirable harms (e.g., financial harms, legal risks) If so, please provide a description. Is there anything a future user could do to mitigate these undesirable harms?}

\textbf{A4:} No.

\noindent \textbf{5. Are there tasks for which the dataset should not be used? If so, please provide a description.}

\textbf{A5:} No.

\subsection{Distribution}

\noindent \textbf{1. Will the dataset be distributed to third parties outside of the entity (e.g., company, institution, organization) on behalf of which the dataset was created? If so, please provide a description.}

\textbf{A1:} Yes. The dataset will be made publicly available to the research community.

\noindent \textbf{2. How will the dataset will be distributed (e.g., tarball on website, API, GitHub)? Does the dataset have a digital object identifier (DOI)?}

\textbf{A2:} It will be publicly available on the project website at \href{https://github.com/pandorgan/APT-36K}{GitHub}.

\noindent \textbf{3. When will the dataset be distributed?}

\textbf{A3:} The dataset will be distributed once the paper is accepted after peer-review.

\noindent \textbf{4. Will the dataset be distributed under a copyright or other intellectual property (IP) license, and/or under applicable terms of use (ToU)? If so, please describe this license and/or ToU, and provide a link or other access point to, or otherwise reproduce, any relevant licensing terms or ToU, as well as any fees associated with these restrictions.}

\textbf{A4:} It will be distributed under the CC-BY-4.0 licence.

\noindent \textbf{5. Have any third parties imposed IP-based or other restrictions on the data associated with the instances? If so, please describe these restrictions, and provide a link or other access point to, or otherwise reproduce, any relevant licensing terms, as well as any fees associated with these restrictions.}

\textbf{A5:} No.

\noindent \textbf{6. Do any export controls or other regulatory restrictions apply to the dataset or to individual instances? If so, please describe these restrictions, and provide a link or other access point to, or otherwise reproduce, any supporting documentation.}

\textbf{A6:} No.

\subsection{Maintenance}

\noindent \textbf{1. Who will be supporting/hosting/maintaining the dataset?}

\textbf{A1:} The authors.

\noindent \textbf{2. How can the owner/curator/manager of the dataset be contacted (e.g., email address)?}

\textbf{A2:} They can be contacted via email available on the project website.

\noindent \textbf{3. Is there an erratum? If so, please provide a link or other access point.}

\textbf{A3:} No. 

\noindent \textbf{4. Will the dataset be updated (e.g., to correct labeling errors, add new instances, delete instances)? If so, please describe how often, by whom, and how updates will be communicated to users (e.g., mailing list, GitHub)?}

\textbf{A4:} No. Since we conduct three-round annotation check, we believe the annotation errors are very seldom in our APT-36K dataset. The occasional mistakes such as annotation drift can be treated as noise in the dataset. 

\noindent \textbf{5. Will older versions of the dataset continue to be supported/hosted/maintained? If so, please describe how. If not, please describe how its obsolescence will be communicated to users.}

\textbf{A5:} N/A.

\noindent \textbf{6. If others want to extend/augment/build on/contribute to the dataset, is there a mechanism for them to do so? If so, please provide a description. Will these contributions be validated/verified? If so, please describe how. If not, why not? Is there a process for communicating/distributing these contributions to other users? If so, please provide a description.}

\textbf{A6:} N/A. 

\small{
\normalem
\bibliographystyle{ieee}
\bibliography{egbib}
}

\end{document}